\documentclass[conference]{IEEEtran}
\IEEEoverridecommandlockouts
\usepackage{cite}
\usepackage{amsmath,amssymb,amsfonts}
\usepackage{algorithmic}
\usepackage{graphicx}
\usepackage{textcomp}
\usepackage{xcolor}
\def\BibTeX{{\rm B\kern-.05em{\sc i\kern-.025em b}\kern-.08em
    T\kern-.1667em\lower.7ex\hbox{E}\kern-.125emX}}
\begin{document}

\title{A Dual-Branch CNN for Robust Detection of AI-Generated Facial Forgeries
}

\author{
\IEEEauthorblockN{1\textsuperscript{st} Xin Zhang}
\IEEEauthorblockA{\textit{Computer Science Department} \\
\textit{University of Southern Maine}\\
Portland, United States \\
xin.zhang@maine.edu}
\and
\IEEEauthorblockN{2\textsuperscript{nd} Yuqi Song}
\IEEEauthorblockA{\textit{Computer Science Department} \\
\textit{University of Southern Maine}\\
Portland, United States \\
yuqi.song@maine.edu}
\and
\IEEEauthorblockN{3\textsuperscript{rd} Fei Zuo}
\IEEEauthorblockA{\textit{The Department of Computer Science} \\
\textit{	University of Central Oklahoma	}\\
Edmond, United States \\
fzuo@uco.edu}
}

\maketitle

\begin{abstract}
The rapid advancement of generative AI has enabled the creation of highly realistic forged facial images, posing significant threats to AI security, digital media integrity, and public trust. Face forgery techniques—ranging from face swapping and attribute editing to powerful diffusion-based image synthesis—are increasingly being used for malicious purposes such as misinformation, identity fraud, and defamation. This growing challenge underscores the urgent need for robust and generalizable face forgery detection methods as a critical component of AI security infrastructure.
In this work, we propose a novel dual-branch convolutional neural network for face forgery detection that leverages complementary cues from both spatial and frequency domains. The RGB branch captures semantic information, while the frequency branch focuses on high-frequency artifacts that are difficult for generative models to suppress. A channel attention module is introduced to adaptively fuse these heterogeneous features, highlighting the most informative channels for forgery discrimination. To guide the network’s learning process, we design a unified loss function—FSC Loss—that combines focal loss, supervised contrastive loss, and a frequency center margin loss to enhance class separability and robustness.
We evaluate our model on the DiFF benchmark, which includes forged images generated from four representative methods: text-to-image, image-to-image, face swap, and face edit. Our method achieves strong performance across all categories and outperforms average human accuracy. These results demonstrate the model’s effectiveness and its potential contribution to safeguarding AI ecosystems against visual forgery attacks.
\end{abstract}


\section{Introduction}
The rapid advancement of generative artificial intelligence (AI), particularly in deep generative models such as GANs and diffusion models, has transformed the way digital images are created~\cite{verdoliva2020media}. These models can synthesize highly realistic, photorealistic images that are virtually indistinguishable from real ones, even to trained observers~\cite{mirsky2021creation}. While these technologies have led to remarkable progress in creative industries, advertising, entertainment, and virtual reality, they also pose significant societal threats when misused for malicious purposes~\cite{chesney2019deep}. Image forgeries can be exploited for identity theft, political propaganda, harassment, and large-scale misinformation campaigns, eroding trust in digital media and compromising personal and national security~\cite{tolosana2020deepfakes}. With the widespread dissemination of forged content on social platforms, the development of \textbf{robust image forgery detection systems} is more critical than ever.

\begin{figure}[h]
    \centering
    \includegraphics[width=0.47\textwidth]{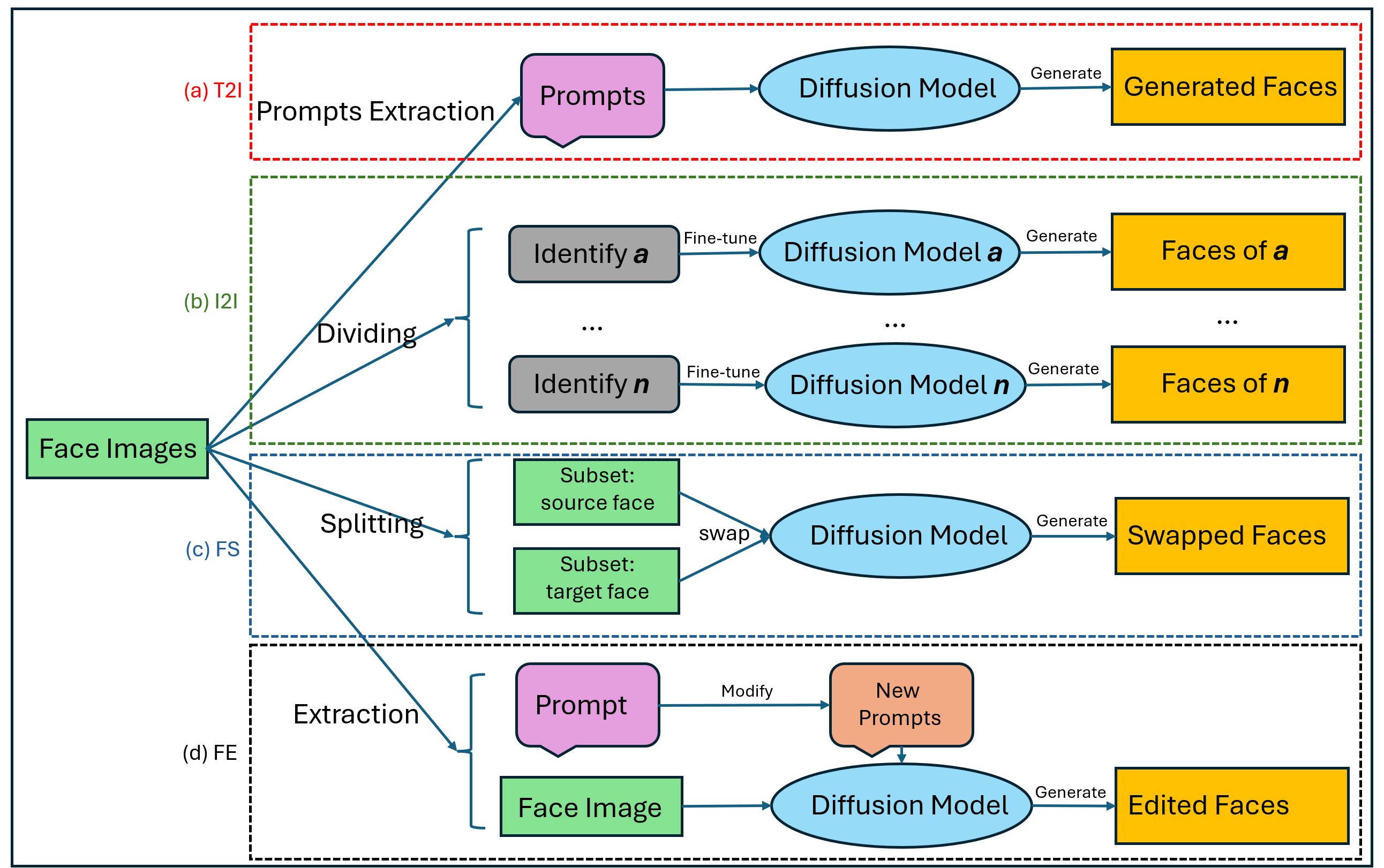}
    \caption{Detailed pipelines for generating diffusion-based facial forgeries under four conditions.
(a) T2I: Prompts are extracted from face images and fed into a diffusion model to synthesize new faces.
(b) I2I: The dataset is divided by identities (e.g., identity $a$ to $n$), and each identity is used to fine-tune a separate diffusion model to generate diverse images of that specific identity.
(c) FS: The dataset is split into source and target subsets; the diffusion model swaps identity features between the two subsets to produce realistic swapped faces.
(d) FE: Prompts and face images are extracted, the prompts are modified, and the updated prompts are passed to a diffusion model to edit attributes such as expression, age, or style while preserving the core facial features.}
    \label{fig:four}
\end{figure}

As illustrated in the DiFF benchmark~\cite{cheng2024diffusion}, modern image forgeries can be generated using four primary methods: (a) text-to-image (T2I) generation from textual prompts, (b) image-to-image (I2I) synthesis via fine-tuned diffusion models, (c) face swapping (FS) where the identity in a target image is replaced with a source face, and (d) face editing (FE) where attributes such as age, expression, or gender are altered using prompt-based modifications. All these techniques leverage advanced generative models that produce forgeries with highly natural textures, making detection extremely challenging. The detailed generation pipelines are shown in Fig.~\ref{fig:four}.

The detection challenge is evident in the DiFF dataset, where human performance is only 59.65\% for text-to-image, 59.63\% for image-to-image, 51.50\% for face swapping, and 45.53\% for face editing~\cite{cheng2024diffusion}. Such results show that even human observers struggle to reliably identify diffusion-based forgeries, highlighting the need for automated detection systems capable of capturing subtle artifacts in both spatial and frequency domains.

In this paper, we propose a dual-branch CNN-based forgery detection model that integrates spatial and frequency information using channel attention fusion. The RGB branch learns semantic and structural features, while the frequency branch focuses on high-frequency residuals where imperceptible generative artifacts are often present. The two branches are fused via a channel attention module, which adaptively assigns weights to the most informative features from each branch. This design enables the network to effectively leverage complementary cues for forgery detection. To further enhance discriminative feature learning, we introduce FSC Loss (Focal–Supervised Contrastive–Center Loss), which consists of three components: focal loss~\cite{lin2017focal}, supervised contrastive loss~\cite{khosla2020supervised}, and frequency center margin loss~\cite{li2021frequency}. The combination is designed to not only emphasize hard-to-classify samples (via focal loss) but also to improve the compactness and separability of feature embeddings (via supervised contrastive loss) while ensuring that frequency-domain features of real and fake images remain well-distinguished (via frequency center margin loss).

Our main contributions are as follows:
\begin{itemize}
\item We propose a dual-branch ResNet-based architecture with channel attention fusion, effectively combining spatial and frequency cues for robust forgery detection.
\item We design a novel FSC Loss, integrating focal, supervised contrastive, and frequency center margin objectives to improve classification performance and feature separability.
\item We conduct extensive experiments on the DiFF dataset, demonstrating that our approach outperforms existing detectors across all four forgery categories (T2I, I2I, FS, FE).
\end{itemize}

\section{Realted Work}
The research on face forgery detection has evolved from early handcrafted forensic features to deep learning-based methods that can capture subtle artifacts introduced by modern generative models. Traditional techniques—such as splicing detection, copy-move detection, and noise inconsistency analysis—relied on pixel-level anomalies like unnatural transitions or lighting inconsistencies~\cite{bayar2016deep,farid2009exposing}. While effective for conventional image tampering, these methods fail on AI-generated forgeries, which exhibit smooth textures and high photorealism.

With the rise of GAN-based deepfakes, CNNs became the foundation of forgery detection. XceptionNet demonstrated strong performance by capturing pixel-level manipulation patterns~\cite{rossler2019faceforensics++, rossler2018faceforensics}, while Face X-ray~\cite{li2020face} improved detection by focusing on blending artifacts near manipulated boundaries. To better generalize to unseen forgeries, frequency-domain methods have been explored. 
It has been shown that GAN-generated images exhibit distinctive anomalies in the Fourier spectrum~\cite{dzanic2020fourier,hu2021exposing}.
F3-Net~\cite{qian2020thinking} combines spatial and frequency cues in a multi-branch architecture, leading to robust performance across manipulation types.

More recently, attention mechanisms~\cite{zhao2021multi} and transformer-based models~\cite{wang2022m2tr} have been applied to focus on manipulation-prone regions and long-range dependencies. However, diffusion models pose a new challenge, as their iterative denoising process produces fewer frequency artifacts and more natural textures~\cite{pashine2021deep,dhariwal2021diffusion}, making traditional GAN-trained detectors less effective. This motivates hybrid approaches which captures both spatial structures and high-frequency residual cues to address diffusion-based forgeries.

\section{Proposed Method}
\begin{figure*}[h]
    \centering
    \includegraphics[width=0.9\textwidth]{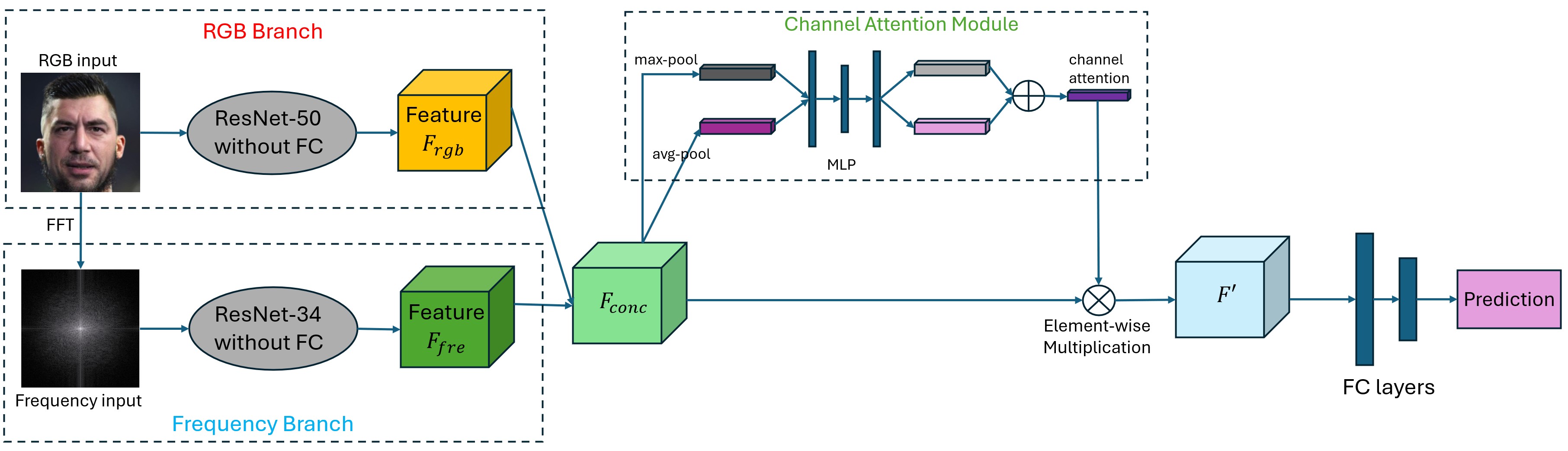}
    \caption{\textbf{Overview of the proposed dual-branch detection framework.}
        The RGB image is processed through a ResNet-50 backbone, while its frequency representation (obtained via FFT) is passed through a ResNet-34. The resulting features are concatenated and refined using a Channel Attention Module, which adaptively emphasizes informative channels. The fused feature is then pooled and passed through fully connected layers for real-vs-fake prediction.}
    \label{fig:pp}
\end{figure*}

The overall pipeline of our proposed face forgery detection model is illustrated in Figure~\ref{fig:pp}. Our framework adopts a dual-branch architecture that leverages both RGB and frequency-domain cues to enhance detection robustness.

First, the input face image is simultaneously processed through two pathways:
\begin{itemize}
\item The RGB branch receives the original image and extracts spatial features using a ResNet-50 backbone (with the classification head removed).
\item The frequency branch first converts the RGB image to the frequency domain using a Fast Fourier Transform (FFT). The resulting spectrum is then passed through a ResNet-34 backbone to capture complementary frequency-based features.
\end{itemize}

The outputs of these two branches are concatenated along the channel dimension to form a unified representation. To adaptively emphasize the most informative channels, we apply a Channel Attention Module to the concatenated features. This module learns to reweight channels based on their relevance to the forgery detection task.

The attention-enhanced feature is then passed through fully connected layers to perform binary classification, predicting whether the input image is real or fake.

\subsection{Dual-branch Design}
To enhance the model’s ability to detect subtle and diverse artifacts introduced by modern generative techniques, we adopt a dual-branch architecture that processes input images through both RGB and frequency domains. This design leverages complementary information: spatial texture patterns in the RGB space and spectral artifacts in the frequency domain, resulting in improved robustness and generalization.

\noindent
\textbf{Motivation for Dual-Branch Design.} Purely spatial (RGB-based) detectors can struggle to identify forgeries generated by high-fidelity generative models such as diffusion-based text-to-image or face-editing systems. These forgeries often exhibit highly natural visual appearances, with little to no low-level anomalies observable in the pixel domain. However, as highlighted by prior works such as~\cite{frank2020leveraging,qian2020thinking}, generative models often leave behind frequency inconsistencies due to non-linear activations, interpolation, and upsampling operations. These spectral anomalies are hard for human observers to detect but can be exploited algorithmically.

By integrating both the RGB and frequency views, the model captures:
\begin{itemize}
\item Local texture patterns and spatial dependencies via the RGB branch.
\item Global periodic and spectral characteristics via the frequency branch.
\end{itemize}

This dual design allows the model to detect a wider range of forgery traces and improves generalizability across manipulation types.

\noindent
\textbf{Frequency Input Generation. }
To capture subtle frequency artifacts introduced by generative models, we transform each RGB image into a frequency-domain representation. This is achieved through the following steps:
\begin{itemize}
\item The RGB image $I_{\text{rgb}} \in \mathbb{R}^{H \times W \times 3}$ is first converted into a single-channel grayscale image $I_{\text{gray}} \in \mathbb{R}^{H \times W}$ using the luminance-preserving weighted sum:

\begin{equation}
I_{\text{gray}} = 0.299 \cdot R + 0.587 \cdot G + 0.114 \cdot B
\end{equation}

where $R$, $G$, and $B$ represent the red, green, and blue channels of the input image, respectively. These coefficients are derived from the ITU-R BT.601 standard and reflect the human eye's varying sensitivity to different wavelengths.
\item We then compute the magnitude spectrum of the grayscale image using the Fast Fourier Transform (FFT)~\cite{brigham1988fast}. To enhance stability and suppress extreme values, we apply logarithmic scaling:

\begin{equation}
I_{\text{fre}}  = \log \left( 1 + |\mathcal{F}(I_{\text{gray}})| \right)
\end{equation}

Here, $\mathcal{F}(\cdot)$ denotes the 2D FFT operator, and $|\cdot|$ represents the magnitude of the complex frequency coefficients. The resulting $I_{\text{fre}} \in \mathbb{R}^{H \times W}$ is used as the input to the frequency branch of our model.
\end{itemize}

\noindent
\textbf{Feature Representation and Fusion.} Given an RGB input image $I \in \mathbb{R}^{3 \times H \times W}$, our model extracts two complementary types of features through a dual-branch architecture:

\begin{itemize}
    \item \textbf{RGB Branch:} The input is processed through a ResNet-50 backbone (excluding the final fully connected layer), resulting in a spatial feature map:
    $$
    F_{\text{rgb}} \in \mathbb{R}^{2048 \times 7 \times 7}
    $$

    \item \textbf{Frequency Branch:} The precomputed frequency representation $I_{\text{fre}}$ is passed through a ResNet-34 backbone to produce:
    $$
    F_{\text{fre}} \in \mathbb{R}^{512 \times 7 \times 7}
    $$
\end{itemize}

The two feature maps are concatenated along the channel dimension and modulated by a channel attention map to highlight informative features. Specifically, we compute:
$$
F' = M_c \odot \text{Concat}(F_{\text{rgb}}, F_{\text{fre}}), \quad M_c \in \mathbb{R}^{2560 \times 1 \times 1}
$$
where $\odot$ stands for the element-wise multiplication, and $F' \in \mathbb{R}^{2560 \times 7 \times 7}$ is the attention-weighted feature map. Global average pooling (GAP) is then applied to obtain the final feature vector:
$$
f = \text{GAP}(F') \in \mathbb{R}^{2560}
$$
which is passed through fully connected layers to produce a binary output indicating whether the image is real or forged.

To compute the channel attention map $M_c$, we follow the strategy of the Convolutional Block Attention Module (CBAM)~\cite{woo2018cbam}. Given the concatenated feature map $F_{\text{conc}}$, we first apply both average pooling and max pooling along the spatial dimensions to generate two descriptors:
$$
f_{\text{avg}} = \text{AvgPool}(F_{\text{conc}}), \quad f_{\text{max}} = \text{MaxPool}(F_{\text{conc}})
$$
Both $f_{\text{avg}}$ and $f_{\text{max}} \in \mathbb{R}^{2560 \times 1 \times 1}$ are passed through a multilayer perceptron (MLP) consisting of two fully connected layers with ReLU activation. The MLP outputs are then summed and passed through a sigmoid function to obtain the final channel attention map:
$$
M_c = \sigma(\text{MLP}(f_{\text{avg}}) + \text{MLP}(f_{\text{max}}))
$$
This attention mechanism adaptively emphasizes the most informative channels for the classification task.

\subsection{The FSC loss function}
To effectively supervise the training process and encourage the network to learn discriminative features across both RGB and frequency domains, we propose a novel loss function called \textbf{FSC Loss}, which integrates three components: Focal Loss~\cite{lin2017focal}, Supervised Contrastive Loss~\cite{khosla2020supervised}, and Frequency Center Margin Loss~\cite{li2021frequency}. The overall loss function is defined as:

$$
\mathcal{L}_{\text{FSC}} = \mathcal{L}_{\text{focal}} + \lambda_1 \cdot \mathcal{L}_{\text{supcon}} + \lambda_2 \cdot \mathcal{L}_{\text{f-center}}
$$

where $ \lambda_1 $ and $ \lambda_2 $ are weighting coefficients that balance the contribution of each term.

\begin{itemize}
    \item \textbf{Focal Loss} ($ \mathcal{L}_{\text{focal}} $). This term addresses class imbalance and emphasizes hard-to-classify samples. It modifies the standard cross-entropy loss with a focusing parameter $ \gamma $ and a weighting factor $ \alpha $. For a binary classification problem with label $ y \in \{0,1\} $ and predicted probability $ p $, the focal loss is:

    $$
    \mathcal{L}_{\text{focal}} =
    \begin{cases}
    - \alpha (1 - p)^\gamma \log(p), & \text{if } y = 1 \\
    - (1 - \alpha) p^\gamma \log(1 - p), & \text{if } y = 0
    \end{cases}
    $$

\item \textbf{Supervised Contrastive Loss} ($ \mathcal{L}_{\text{supcon}} $). This loss improves feature representation by encouraging embeddings from the same class to be close together, and those from different classes to be far apart. Given a batch of samples, for each anchor feature $ z_i $, we consider the set of positive features $ P(i) $ (same class) and contrast them against all other features $ A(i) $ (excluding $ i $ itself). The loss is:

$$
\mathcal{L}_{\text{supcon}} = \sum_{i \in I} \frac{-1}{|P(i)|} \sum_{p \in P(i)} \log \frac{\exp(z_i \cdot z_p / \tau)}{\sum_{a \in A(i)} \exp(z_i \cdot z_a / \tau)}
$$

Here, $ \cdot $ denotes dot product similarity, $ \tau $ is a temperature parameter controlling the sharpness of the distribution, and $ z_i $ is treated as the anchor feature. This helps create compact clusters for each class in the embedding space.

\item \textbf{Frequency Center Margin Loss} ($ \mathcal{L}_{\text{f-center}} $). This term improves class separation in the frequency feature space. It consists of two parts:
(1) pulling frequency features $ f_i $ closer to their corresponding class centers $ c_{y_i} $, and
(2) pushing the centers of different classes away from each other by at least a fixed margin $ m $. The loss is formulated as:

$$
\mathcal{L}_{\text{f-center}} = \sum_{i=1}^N \|f_i - c_{y_i}\|_2^2 + \mu \sum_{j \neq k} \max(0, m - \|c_j - c_k\|_2)^2
$$

The first term ensures that features of the same class are compact, while the second term penalizes class centers that are too close. This helps the model distinguish between real and fake images based on frequency-domain cues.
 
\end{itemize}

By combining these three complementary components, the FSC loss enables the model to learn robust and semantically meaningful features from both spatial and spectral perspectives.

\section{Experiments}

\subsection{Dataset}
We evaluate our method using the DiFF benchmark~\cite{cheng2024diffusion}, a recently introduced dataset dedicated to facial image forgery detection. DiFF is designed to reflect real-world forgery generation techniques by incorporating four representative manipulation categories: Text-to-Image (T2I), Image-to-Image (I2I), Face Swap (FS), and Face Edit (FE). These categories span a wide range of generative pipelines used in practice.

\begin{itemize}
    \item \textbf{Text-to-Image (T2I):} Images are synthesized from prompts using pretrained diffusion models such as FreeDoM~\cite{yu2023freedom}.
    \item \textbf{Image-to-Image (I2I):} Images are generated from real faces via fine-tuned diffusion models like LoRA~\cite{hu2022lora}, preserving general structure but altering identity or expression.
    \item \textbf{Face Swap (FS):} Identity information from one face is transplanted onto another using methods like DiffFace~\cite{kim2025diffface}.
    \item \textbf{Face Edit (FE):} Attribute-level changes (e.g., age, expression, gender) are applied using prompt-based image editing techniques such as CoDiff~\cite{huang2023collaborative} and Imagic~\cite{kawar2023imagic}.
\end{itemize}

For each category, DiFF includes both pristine (real) and manipulated (fake) samples in a 1:1 ratio. The dataset is curated to avoid identity leakage between training and test sets by following an identity-disjoint protocol. Evaluation splits are provided to support both in-domain and cross-domain benchmarking, enabling robust assessment of model generalization across forgery types.

We select the DiFF benchmark for its diversity, realism, and coverage of modern forgery techniques.Unlike earlier datasets that focus only on GAN-based manipulations, DiFF comprehensively evaluates detection performance across diffusion-generated and edited content, better reflecting today’s forgery landscape.

\subsection{Experimental Results}

\noindent
\textbf{In-domain Performance.} In-domain performance refers to the setting where the model is trained and evaluated on the same type of forgery, allowing it to specialize in learning manipulation-specific features. This evaluation serves as a fundamental benchmark to assess the model’s upper-bound performance and its ability to capture intrinsic patterns within each forgery category, prior to testing its generalization to unseen domains. As summarized in Table~\ref{tab:in-domain-performance}, our proposed method achieves strong in-domain detection results across all forgery types, performing comparably to, and in some cases outperforming, state-of-the-art baselines. These results demonstrate that the dual-branch architecture, combined with the FSC loss, effectively leverages both spatial and frequency-domain cues to identify forgeries with high accuracy when trained and tested within the same manipulation domain. This highlights the model’s ability to learn highly discriminative representations under in-distribution conditions.

\begin{table}[ht]
\centering
\caption{In-domain detection accuracy (\%) on the DiFF benchmark. Each model is trained and evaluated on the same forgery type using an 80/20 train-test split. The best-performing method is highlighted in bold.}
\label{tab:in-domain-performance}
\begin{tabular}{|l|c|c|c|c|}
\hline
\textbf{Method} & \textbf{T2I-T2I} & \textbf{I2I-I2I} & \textbf{FS-FS} & \textbf{FE-FE} \\ \hline
Xception~\cite{rossler2019faceforensics++}      & 93.32 & 98.92 & 99.95 & \textbf{99.95} \\ \hline
F$^3$-Net~\cite{qian2020thinking}      & 99.60 & 99.50 & \textbf{99.98} & 99.60 \\ \hline
EfficientNet~\cite{tan2019efficientnet}   & 99.89 & 99.80 & 99.87 & 99.24 \\ \hline
DIRE~\cite{wang2023dire}           & 95.04 & 99.88 & 99.09 & 99.87 \\ \hline
\textbf{Ours}  & \textbf{99.91} & \textbf{99.89} & 99.94 & 99.07  \\\hline
\end{tabular}
\end{table}

\noindent
\textbf{Cross-domain Performance.} Cross-domain evaluation assesses the model's ability to generalize across different types of forgery generation methods. Specifically, models are trained on one subset (e.g., T2I) and tested on other subsets (e.g., I2I, FS, FE). This setting reflects real-world scenarios where the type of forgery encountered during deployment may differ from that seen during training.
As shown in Table.~\ref{tab:cross_generalization}, our model exhibits strong generalization ability across different forgery generation methods. When trained on the T2I subset, it achieves the best performance on all other test sets: 90.21\% on I2I, 47.98\% on FS, and 73.70\% on FE. Under the I2I training setting, our model outperforms baselines on T2I (91.17\%), FS (47.91\% vs. 41.51\% for DIRE), and FE (50.21\% vs. 46.19\% for F$^3$-Net ). When trained on FS, it again achieves the best performance across T2I (42.30\%), I2I (39.21\%), and FE (36.88\%), substantially surpassing the strongest baseline DIRE. Lastly, under the FE training condition, our method reaches the highest score on T2I (85.32\%), and performs competitively on I2I (78.34\% vs. 79.12\%) and FS (68.97\% vs. 70.81\%). These results consistently demonstrate our model’s robustness in cross-domain scenarios, where training and testing are conducted on different manipulation techniques.

\begin{table}[t]
\centering
\caption{Cross-domain generalization results. Each model is trained on one subset and evaluated on all subsets. The best results are highlighted in bold, while the second-best results are underlined.}
\label{tab:cross_generalization}
\begin{tabular}{|l|c|c|c|c|c|}
\hline
\textbf{Methods} & \textbf{Train Subset} &  \multicolumn{4}{c|}{\textbf{Testing Set}} \\
\cline{3-6}
 & & \textbf{T2I} & \textbf{I2I} & \textbf{FS} & \textbf{FE} \\
\hline
Xception     & T2I & - & 86.85 & 34.65 & 23.28 \\ \hline
F$^3$-Net    & T2I & - & 88.50 & \underline{45.07} & \underline{71.06} \\ \hline
EfficientNet & T2I & - & \underline{89.72} & 21.49 & 49.63 \\ \hline
DIRE         & T2I & - & 84.07 & 35.15 & 50.86 \\ \hline
Ours         & T2I & - & \textbf{90.21} & \textbf{47.98} & \textbf{73.70} \\ \hline
\hline
Xception     & I2I & \underline{87.82} & - & 36.82 & 33.39 \\ \hline
F$^3$-Net    & I2I & 87.23 & - & 40.62 & \underline{46.19} \\ \hline
EfficientNet & I2I & 84.39 & - & 19.47 & 27.46 \\ \hline
DIRE         & I2I & 86.20 & - & \underline{41.51} & 42.01 \\ \hline
Ours         & I2I & \textbf{91.17} & - & \textbf{47.91} & \textbf{50.21} \\ \hline
\hline
Xception     & FS  & 23.17 & 24.47 & - & 10.17 \\ \hline
F$^3$-Net    & FS  & \underline{35.43} & 30.39 & - & 20.79 \\ \hline
EfficientNet & FS  & 16.88 & 22.17 & - & 10.21 \\ \hline
DIRE         & FS  & 16.08 & \underline{36.27} & - & \underline{32.68} \\ \hline
Ours         & FS  & \textbf{42.30} & \textbf{39.21} & - & \textbf{36.88} \\ \hline
\hline
Xception     & FE  & 80.84 & \textbf{79.12} & \textbf{70.81} & - \\ \hline
F$^3$-Net    & FE  & \underline{82.32} & 76.92 & 56.27 & - \\ \hline
EfficientNet & FE  & 80.41 & 63.06 & 66.62 & - \\ \hline
DIRE         & FE  & 56.70 & 59.22 & 43.78 & - \\ \hline
Ours         & FE  & \textbf{85.32} & \underline{78.34} & \underline{68.97} & - \\
\hline
\end{tabular}
\end{table}

\noindent
\textbf{Comparison with Human Performance.} Table~\ref{tab:human_comparison} presents a comparison between human performance and our model across four forgery generation categories. The second to fifth columns correspond to detection performance on each type of forgery: T2I (text-to-image), I2I (image-to-image), FS (face swap), and FE (face edit).
Human performance values are taken from the DiFF benchmark~\cite{cheng2024diffusion}, which conducted a large-scale user study involving 70 participants. Each participant was instructed to classify the authenticity of 200 randomly selected images with a 50:50 split between real and fake images, ensuring no identity repetition to avoid bias. In total, 14,000 human classification decisions were collected. We report the average accuracy per forgery category.
Our model’s results are computed by averaging the detection accuracies for each category across \textbf{all corresponding cross-domain test settings}. The comparison shows that our method consistently outperforms human observers in all categories. These results demonstrate the effectiveness of our model in detecting highly realistic forgeries that are often difficult even for humans to identify.

\begin{table}[h]
\centering
\caption{Comparison with human performance across different forgery types.}
\begin{tabular}{|l|c|c|c|c|}
\hline
\textbf{Method} & \textbf{T2I} & \textbf{I2I} & \textbf{FS} & \textbf{FE} \\
\hline
Human & 59.65 & 59.63 & 51.50 & 45.53 \\ \hline
Ours & \textbf{72.92} & \textbf{69.25} & \textbf{54.95} & \textbf{53.60} \\
\hline
\end{tabular}
\label{tab:human_comparison}
\end{table}

\noindent 
\textbf{Ablation Study.} To evaluate the effectiveness of individual components in our architecture, we conduct ablation experiments under three configurations:
\begin{itemize}
    \item w/o \textit{Fre-Branch}: Removes the frequency branch, using only the RGB stream for feature extraction.

    \item w/o $\mathcal{L}_{\text{f-center}}$: Excludes the frequency center margin loss from the training objective.

    \item w/o $M_c$: Eliminates the channel attention mechanism; features from both branches are directly concatenated without attention weighting.
\end{itemize}

As shown in Table.~\ref{tab:as}, removing the frequency branch leads to the most significant drop in performance, with an average decrease of 13.4\% across test sets, demonstrating the crucial role of frequency information.
Eliminating $\mathcal{L}_{\text{f-center}}$ causes noticeable drops as well, particularly on the FE set (6.79\%$\downarrow$), confirming that frequency-specific supervision strengthens discriminability.
Omitting $M_c$ results in degraded performance on all subsets, with a 3.1\% drop on average, indicating the importance of attention-guided feature fusion.
These results highlight that each component (frequency representation, frequency-aware loss, and attention fusion) contributes meaningfully to the model’s overall performance.

\begin{table}[t]
\centering
\caption{Ablation Study of Key Components. This table quantifies the impact of removing individual components of our model: the frequency branch, the frequency center loss ($\mathcal{L}_{\text{f-center}}$), and the channel attention module ($M_c$). The performance drop across subsets highlights the effectiveness of each design.}
\label{tab:as}
\begin{tabular}{|l|c|c|c|c|}
\hline
\textbf{Configurations} &  \multicolumn{4}{c|}{\textbf{Testing Set}} \\
\cline{2-5}
 & \textbf{T2I} & \textbf{I2I} & \textbf{FS} & \textbf{FE} \\
\hline
w/o \textbf{Fre-Branch} & 94.23 & 82.31 & 33.68 & 42.78  \\\hline
w/o $ \mathcal{L}_{\text{f-center}} $ & 99.06 & 88.17 & 43.01 & 66.91   \\\hline
w/o $M_c$ & 97.45 & 88.29 & 45.88 &  67.74 \\\hline
Ours         & \textbf{99.91} & \textbf{90.21} & \textbf{47.98} & \textbf{73.70} \\ \hline

\end{tabular}
\end{table}

\section{Conclusion}
With the rapid evolution of generative models, forged facial images have become increasingly realistic and widespread, posing serious threats to digital media integrity, public trust, and security. Detecting such forgeries—especially those generated by advanced diffusion-based methods—is essential for safeguarding the authenticity of visual content in the AI era.

To address this challenge, we propose a novel detection framework capable of handling a wide range of forgery generation techniques, including text-to-image synthesis, image-to-image translation, face swapping, and face editing. Our method leverages a dual-branch CNN architecture that captures complementary cues from both RGB and frequency domains, fused via a channel attention mechanism to enhance informative features. Additionally, we introduce FSC Loss, a composite objective combining focal loss, supervised contrastive loss, and frequency center margin loss, to promote discriminative and robust feature learning.

Comprehensive experiments on the DiFF benchmark demonstrate the effectiveness of our approach. The model consistently performs well across both in-domain and cross-domain settings, and significantly outperforms average human detection accuracy. These results highlight the importance and feasibility of building generalized and trustworthy forgery detection systems for real-world applications.

\bibliographystyle{IEEEtran}
\bibliography{ref}

\end{document}